\documentclass{article}

\usepackage{PRIMEarxiv}

\usepackage[utf8]{inputenc} 
\usepackage[T1]{fontenc}    
\usepackage{hyperref}       
\usepackage{url}            
\usepackage{booktabs}       
\usepackage{amsfonts}       
\usepackage{nicefrac}       
\usepackage{microtype}      
\usepackage{lipsum}
\usepackage{fancyhdr}       
\usepackage{graphicx}       
\graphicspath{{media/}}     
\usepackage{bbding}
\usepackage{amsmath}

\title{CT-block: a novel local and global features extractor for point cloud
}

\author{
  Shangwei Guo, Jun Li, Zhengchao Lai, Xiantong Meng, Shaokun Han* \\
  Beijing Key Lab for Precision Optoelectronic Measurement Instrument and Technology\\
  Beijing Institute of Technology, School of Optics and Photonics  \\
  Beijing 100081, China\\
  \texttt{skhan@bit.edu.cn(* indicates the corresponding author)} \\
}

\begin{document}
\maketitle

\begin{abstract}
Deep learning on the point cloud is increasingly developing. Grouping the point with its neighbors and conducting convolution-like operation on them can learn the local feature of the point cloud, but this method is weak to extract the long-distance global feature. Performing the attention-based transformer on the whole point cloud can effectively learn the global feature of it, but this method is hardly to extract the local detailed feature. In this paper, we propose a novel module that can simultaneously extract and fuse local and global features, which is named as CT-block. The CT-block is composed of two branches, where the letter C represents the convolution-branch and the letter T represents the transformer-branch. The convolution-branch performs convolution on the grouped neighbor points to extract the local feature. Meanwhile, the transformer-branch performs offset-attention process on the whole point cloud to extract the global feature. Through the bridge constructed by the feature transmission element in the CT-block, the local and global features guide each other during learning and are fused effectively. We apply the CT-block to construct point cloud classification and segmentation networks, and evaluate the performance of them by several public datasets. The experimental results show that, because the features learned by CT-block are much expressive, the performance of the networks constructed by the CT-block on the point cloud classification and segmentation tasks achieve state of the art.

\end{abstract}


\section{Introduction}
\label{section1}
Point cloud is a raw data format that can directly express the geometric information of a three-dimensional(3D) object. With the rapid development of 3D scanner, such as LiDAR, binocular camera, and depth camera, etc., the point cloud data is easily accessible. Meanwhile, the analysis and understanding of the point cloud is of great significance to prompt the development of autonomous driving, robotics and remote sensing\cite{chen2017multi,liang2019multi,jiang2018pointsift}.

Deep learning on vision has developed rapidly in recent years. Performing convolution on the regularly girded image pixel can effectively learn expressive features. Different from image which is pixelated, point clouds are unordered and do not conform to the regular grids as in image. Therefore, how to learn expressive features from unordered point clouds has always been the focus of research.

Some methods first convert the unordered point clouds into the regular data format and then perform traditional convolution on them. For example, the multiple-views-based methods first project the point cloud into images from different perspectives\cite{su2015multi,yu2018multi,wei2020view}, and then apply traditional visual Convolutional-Neural-Network(CNN) to learn the feature of projected images to understand the point cloud. This kind of method can only achieve coarse-grained point cloud target classification task, but cannot be competent for fine-grained point cloud segmentation task. The voxel-based methods first convert the point cloud into the regular 3D voxel\cite{wu20153d,maturana2015voxnet}, and then perform 3D-convolution on the voxel to learn the feature of the point cloud. This kind of method is hard to handle dense point clouds, because the computation and memory footprint of it grow cubically with the resolution. Although the computation burden can be alleviated by the sparse convolution\cite{graham20183d}, the procession of voxelizing will bring information loss.

PointNet\cite{qi2017pointnet}, as a pioneer that directly consumes the unordered point clouds, uses the Multiple-Layer-Perceptron (MLP) and the symmetric function to extract the feature of point cloud. PointNet embeds each single point to the latent through the MLPs, and is not able to extract the local features. PointNet++\cite{qi2017pointnet++} performs convolution-like operation on the grouped neighbor points, which makes up for the shortcoming of PointNet that cannot extract the local feature. Some graph-based methods first construct connected graphs on the grouped neighbor points\cite{simonovsky2017dynamic,wang2019dynamic,liu2019dynamic,shen2018mining,chen2019clusternet}, and then graph-convolution is performed on the locally connected graph to extract the local features of the point cloud. Lastly, the graph-pooling is conducted to coarse the point cloud. Similar to the visual-CNN, PointNet++ and some graph-based methods construct a hierarchical network to extract local features of the point cloud layer by layer. Therefore, these methods inherit the shortcoming from visual-CNN, that is, they cannot effectively extract long-distance global features.

The transformer architecture based on the self-attention mechanism emerged from Natural-Language-Processing(NLP), and it has also shined in the visual field in recent years. The network constructed by transformer architecture has excellent performance on the image classification and recognition tasks\cite{beal2020toward,carion2020end,dosovitskiy2020image,touvron2021training,wu2020visual,yuan2021tokens,zhu2020deformable}.  
For any two input features, the self-attention operation first converts them into \textit{query}, \textit{key}, and \textit{value} space. Then, based on the similarity (dot-product) between the two conversion features on the \textit{query} and \textit{key} space, the weight for feature aggregation on the \textit{value} space is obtained. The above operations are performed between any two of the input features, no matter how far away the positions of the two features are on the underlying Euclidean space, and therefore the transformer architecture can extract the long-distance global features. In addition, the transformer architecture processes the input data in a parallelizable and order-independent way, which is suitable for processing unordered data\cite{guo2021pct}. Therefore, many learning-based point cloud methods apply the transformer as feature extractor. The Point Cloud Transformer(PCT)\cite{guo2021pct} proposed the offset-attention-based transformer architecture as the global feature extractor, which can effectively learn the global features of the point cloud. The Ref\cite{zhao2021point} proposed the Point Transformer network, where the self-attention-based transformer is act on the grouped neighbor points. The transformer module in the Point Transformer Network imitates the local convolution kernel in visual field to extract local features of the point cloud.

In vision field, the transformer architecture is a more generalized feature extractor, which lacks the local prior\cite{dosovitskiy2020image}, and therefore it cannot effectively extract local features. The convolution operation is strong for extracting local detailed features but weak for long-distance global features, while the transformer operation is strong for extracting global features but weak for local features. How to simultaneously extract the local and global features effectively is of great significance for a better understanding of images\cite{peng2021conformer}. Some methods, such as Detr\cite{carion2020end}, first use CNN to extract local features, and then input them into the transformer to learn global feature. In the point cloud field, PCT adopts the above-idea\cite{guo2021pct}, where first extracting local features from the grouped neighbor points and then inputting the local features into the offset-attention-based transformer module to learn the global features. However, this fragmented method of extracting local features first and then global features is obviously not optimal, because local features and global features should not be independent and the learning process of them should guide each other. How to simultaneously extract and fuse the local and global features of the point cloud is of great significance to the understanding of it.

Performing convolution-like operations on the grouped neighbor points can extract the local detailed features, while performing the attention-based transformer operation on the entire point cloud can extract the long-distance global features. Inspired by the article \cite{peng2021conformer} in the visual field, we propose a module CT-block that can simultaneously extract and fuse local and global features of the point cloud in this paper. CT-block consists of two branches, where the letter C represents convolution-branch and the letter T represents transformer-branch. The convolution-branch follows the design of PointNet++ to extract the local feature. Transformer-branch follows the design of offset-attention mechanism to extract the global feature. If the CT-block consists of two non-interfering and independent branches, it will not be able to complete the mutual guidance for learning the local and global features and cannot fuse them effectively.
Therefore, we add two feature transmission elements to the CT-block. The first feature transmission element up-samples the local information extracted by convolution-branch and passes it to the transformer-branch, and the second one down-samples the global information extracted by the transformer-branch and passes it to convolution-branch. Through the feature transmission element, the two branches guide each one during extracting features, and they can fuse the learned feature effectively.

We construct a point cloud classification network and a segmentation network by the proposed CT-block and conduct experiments on several benchmark datasets, such as point cloud classification, shape part segmentation, 
to compare the performance of the proposed network constructed with other state of the art methods.

The main contributions of this paper are as follows:
\begin{itemize}
	\item We design a module CT-block which can simultaneously extract and fuse local and global features of the point cloud. The two branches of CT-block respectively extract the local and global features and guide each other during learning through the feature fusion element. The proposed CT-block can serve as the basic element to the construct the network extracting the feature of the point cloud.
	\item Based on the proposed CT-block, we construct the high-performance point cloud classification network and segmentation network. These networks can be the backbone to analyze and understand the point cloud.
	\item We organize multiple experiments on the point cloud classification and  segmentation tasks to verify that the network constructed by the proposed CT-block can achieve or outperform the performance of some state of the art methods.
\end{itemize}

\section{Related Works}
\label{section2}
\subsection{Deep learning on the point cloud}
\label{section2.1}
\subsubsection{Multiple-views-based and voxel-based method}
\label{section2.1.1}
Different from the 2D image relying on regular pixel gird, the point cloud is a set of sparse unordered 3D points. Such unordered data makes it hard to apply traditional deep learning framework. Some researches first convert the unordered point cloud into an ordered data format. 
The multiple-views-based methods first project the point cloud into 2D images from multiple different perspectives\cite{su2015multi,yu2018multi,wei2020view}, and then apply a shared 2D-CNN to learn the features of a set of projected 2D images. The key of these methods is to fuse a set of features extracted from different perspective images into a single feature to express the point cloud target. Such methods can only achieve the coarse-grained point cloud classification task, but is hard to perform the fine-grained point cloud segmentation task. The voxel-based methods first voxelize the unordered point cloud, and then apply 3D-CNN to learn the features of voxel to understand the point cloud\cite{wu20153d,maturana2015voxnet,graham20183d}. When converting the point cloud to the 3D voxel, a low-resolution voxel will lead to the loss of details, while a high-resolution voxel will bring a huge computational burden\cite{guo2020deep}.
\subsubsection{Point-based method}
\label{section2.1.2}
PointNet\cite{qi2017pointnet}, as the pioneer that directly consumes the unordered point cloud, uses simple MLPs to perform feature transformation on each point, and apply symmetric functions (such as maximum pooling or average pooling) to eliminate the effect of the input order of points. However, this method is not being able to extract the local feature due to the feature embedding is only performed on each point. PointNet++\cite{qi2017pointnet++} is stacked by the proposed sampling layers, grouping layers and PointNet layers, where the sampling layer apply the Furthest Point Sampling(FPS) algorithm to down-sample the point cloud, the grouping layer groups neighbor points based on the K-Nearest-Neighbor(k-NN) algorithm, and the PointNet layer performs feature transformation on each grouped point set. Combing the proposed sampling, grouping and PointNet layer, the PointNet++ performs convolution-like operation on the grouped neighbor points to extract the local features. In addition, many methods directly perform the convolution operation on the 3D point cloud to learn the local features. PointCov\cite{wu2019pointconv} directly defines the convolutional kernels in 3D space, where the kernel consists of a weighting function, which is learned by MLP layers, and a density function, which is learned by an MLP layer and a discretized point density estimation. PointCNN\cite{li2018pointcnn} utilizes the proposed $\mathcal{X}$-conv transformation, which transforms the input points into a latent and the potentially canonical order. Then, the PointCNN applies traditional convolution operator on the ordered transformed features. The Kernel Point convolution(KPconv)\cite{thomas2019kpconv} takes grouped neighbor points as input and processes them with weights spatially located by a set of learnable kernel points. The Position Adaptive convolution(PAconv)\cite{xu2021paconv}, which is the weighted sum of stored weight matrices where the coefficients of these weight matrices are self-adaptively learned from point positions through MLPs, is proposed to learn the local feature of the point cloud.

\subsubsection{Graph-based method}
\label{section2.1.3}
Some graph-based methods extract the local feature of the point cloud by constructing a hierarchical network\cite{simonovsky2017dynamic,wang2019dynamic,liu2019dynamic,shen2018mining,chen2019clusternet}.  The Ref \cite{simonovsky2017dynamic} regards each point as a graph vertex and then connects the vertex to its neighbor points to construct a locally connected graph. The proposed Edge-Condition Convolution (ECC) is applied to learn the features on the local graph, followed by the maximum pooling layer that aggregates the features from neighbors and performs graph coarsening. In DGCNN\cite{wang2019dynamic}, the graph convolution operator is processed on the dynamic updated local graph structure, which is constructed in the feature space, to extract the local feature. The Dynamic Points Agglomeration Module (DPAM)\cite{liu2019dynamic} is proposed based on the graph convolution and can implement the function of sampling, grouping, and pooling. It can dynamically exploit the relation of points and agglomerate points in the semantic space. The local feature of the point cloud can be extracted by the hierarchical network stacked by DPAMs. The KCNet\cite{shen2018mining} learns local features by calculating the affinity between the kernel and the given grouped neighbor points, where the kernel is defined by a set of learnable points characterizing geometric types of local structures. The ClusterNet\cite{chen2019clusternet} proposed a rigorously rotation-invariant module to learn the rotation-invariant feature from grouped neighbor points.

The above point-based method and graph-based method are both imitating the visual CNN to construct a hierarchical network structure, where each layer learns local features and the deeper layer with larger receptive field extracts long-distance global features. However, because of imitating the visual CNN, these methods also inherit weakness from it, that is, the ability to extract local features is strong while that to extract global features is poor\cite{dosovitskiy2020image}. The CT-block we propose in this paper combines the strong ability to extract local features of convolution operation and the strong ability to extract global features of transformer architecture. Therefore, the CT-block can simultaneously learn the local and global features of the point cloud.

\subsection{Transformer}
\label{section2.2}
\subsubsection{Transformer on vision}
\label{section2.2.1}
The network constructed by the attention-based transform architecture has been extensively developed in the visual field in recent years. The ViT\cite{dosovitskiy2020image} proved that a network purely constructed by transformers can perform better than the CNN on visual tasks. The success of the transformer constructed networks in the coarse-grained image classification task\cite{beal2020toward,carion2020end,dosovitskiy2020image,touvron2021training,wu2020visual,yuan2021tokens,zhu2020deformable} and the fine-grained image segmentation task\cite{zheng2021rethinking} also proved that the transformer is suitable for extracting global features. However, due to the lack of introduction of local prior, self-attention-based transformer is weak to extract local detailed feature. Some researches, such as DETR and DeiT\cite{carion2020end,touvron2021training}, apply the CNN to extract local features first and then input the extracted local features as tokens to the transformer to makes up for the poor local feature extraction ability of the self-attention operation.

Different from the above methods, the Ref\cite{peng2021conformer} proposed a network named Conformer to simultaneously extract and fuse the local and global features of the image. The two branches of the Conformer can extract local features and global features separately, and the fusion of the two extracted features is implemented through the feature coupling unit. The experimental results show that, compared with the method that only extracts local or global features and the method that extracts local features first and then extracts global features, the Conformer can simultaneously extract and fuse local and global features, and therefore the features learned by it are more expressive.

\subsubsection{Transformer on the point cloud}
\label{section2.2.2}
The transformer architecture has also been applied on the feature learning of the point cloud. The Point Transformer network\cite{zhao2021point} performs the self-attention-based transformer on the grouped neighbor points, where the transformer is act as the local feature extractor. Therefore, the network can be regarded as a special Graph Attention Network (GAT)\cite{velivckovic2017graph}. The PCT\cite{guo2021pct} extracts local features from the grouped neighbor points first, and then input the extracted local features into the proposed offset-attention transformer to learn global features. The proposed offset-attention transformer inherits the strong global feature extraction ability from the attention operation, and the way that inputting the extracted local features to the transformer makes up for the poor local feature extraction ability of the transformer block. However, this fragmented process of extracting local features first and then global features is obviously not optimal. In the visual field, the Conformer\cite{peng2021conformer} has proved that the way that simultaneously extracting local and global features and making the two kinds of features guide each other during learning is definitely better than the way that separately extracting local features first and then global features. Therefore, in the point cloud field, we should also simultaneously extract the local and global features of the point cloud, and make the two features guide each other during the learning process. Inspired by this idea, we propose the CT-block in this paper.

\section{Method}
\label{section3}
In Section \ref{section3.1} , we briefly introduce the idea of CT-block. In Section \ref{section3.2}, we describe in detail that the two branches and feature transmission elements that make up the CT-block, and how the CT-block extracts and fuses the local and global features. In Section \ref{section3.3}, we construct the point cloud classification and segmentation network respectively by the proposed CT-block.
\subsection{Overall}
\label{section3.1}

The local feature of the point cloud is obtained by embedding the grouped neighbor points composed of a center point and its neighbors into the high-dimensional feature space, and it describes the details of the local area but cannot abstract the long-distance information. The global feature of the point cloud is to embed the entire point cloud into the high-dimensional feature space, and it describes the global information of the point cloud but lacks local details. Therefore, local and global features are equally important and complementary. As the fact in the visual field that extracting and fusing the local and global features of the image can better understand it than only learning one kind of feature\cite{peng2021conformer}, effectively extracting and fusing the local and global features of the point cloud is also of great significance to analyze and understand the point cloud.

To extract and fuse the local and global features of the point cloud simultaneously, we design a novel point cloud feature extraction module composed of two interactive branches, which is CT-block as shown in Fig.\,\ref{figure1}. The letter C represents the convolution-branch, and the letter T represents the transformer-branch. For the convolution-branch, the local feature of grouped neighbor points is extracted through stacked MLPs. For the transformer-branch, the global feature of the entire point cloud is learned by attention-based transformer architecture. The local feature and the global feature of the point cloud are complementary. If the two branches are independent and there is no interaction, it will be equivalent to extracting the local feature and the global feature separately from the two branches. Two independent branches cannot effectively make the two kinds of features guide each other during feature learning, and cannot integrate local and global features. Therefore, we design two feature transmission elements between the two branches to make them communicate with each other. The first transmission fusion element up-samples the local feature learned by the convolution-branch and passes it to the transformer-branch. The second feature transmission element down-samples the global feature learned by the transformer-branch and passes it to the convolution-branch. Through the feature transmission element, CT-block can interactively fuse the learned local and global features to extract more expressive features and make them guide each other during learning.

Applying the CT-block as the feature extractor, we can construct the point cloud encoder by stacking it layer by layer. For the $i$-th layer CT-block, its input is the local features $\mathbf{F}_{l}^{i-1}$ and the global features $\mathbf{F}_{g}^{i-1}$ output from the two branches of the ($i$-1)-th layer CT-block. The output of $i$-th layer CT-block is determined by the following equations

\begin{equation}
\begin{aligned}
&\mathbf{F}_{l}^{i}=conv_{2}^{i}\left( conv_{1}^{i}\left( \mathbf{F}_{l}^{i-1} \right) +ft_{2}^{i}\left( \mathbf{F}_{g}^{i} \right) \right)\\
&\mathbf{F}_{g}^{i}=trans\left( ft_{1}^{i}\left( conv_{1}^{i}\left( \mathbf{F}_{l}^{i-1} \right) \right) +\mathbf{F}_{g}^{i-1} \right)
\end{aligned}
\label{formular1}
\end{equation}
where $\mathbf{F}_{l}^{i}$ is the output local feature and $\mathbf{F}_{g}^{i}$ is the output global feature; $conv_{1}^{i}\left( \odot \right) $ represents the first convolution process of the convolution-branch and $conv_{2}^{i}\left( \odot \right) $ is the second of it; $trans\left( \odot \right) $ represents the attention-operation of the transformer-branch; $ft_{1}^{i}\left( \odot \right) $ represents the operation of the first feature transmission element, that is, the local feature extracted by the convolution-branch is passed to the transformer-branch; $ft_{2}^{i}\left( \odot \right) $ represents the operation of the second feature transmission element, that is, the global feature extracted by the transformer-branch is passed to the convolution-branch.

The implementation details of the above processing will be described in Section \ref{section3.2}. From Eq.\,(\ref{formular1}), we can find that the transformer-branch extracting the global features can obtain the local features learned by the convolution-branch through the first feature transmission element, and the convolution-branch extracting the local features can obtain the global features learned by the transformer-branch through the second feature transmission element. The local and global features extracted by CT-block can be learned by mutual guidance and fused effectively.

\subsection{CT-block}
\label{section3.2}
\begin{figure}[ht!]
	\centering\includegraphics[width=16cm]{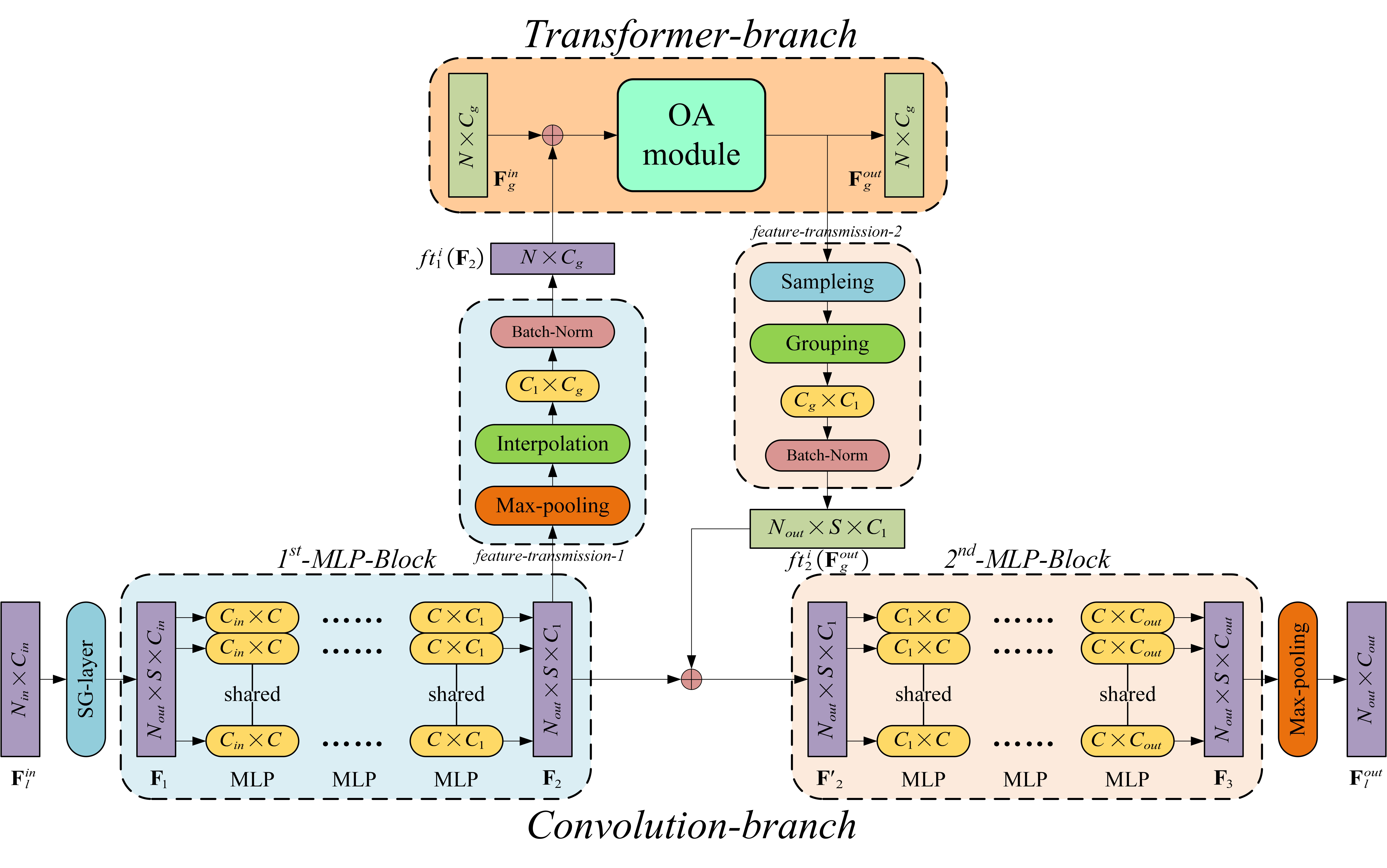}
	\caption{The structure of CT-block. The CT-block consists of a convolution-branch, a transformer-branch, and two feature transmission elements. The convolution-branch extracts the local detailed features of the point cloud. The transformer-branch extracts long-distance global features of the point cloud. The feature transmission element as the bridge makes two branches to guide each other and fuse the features during learning.}
	\label{figure1}
	\vspace{0cm}
\end{figure}

\textbf{Convolution-branch} embeds grouped neighbor points to local features through the stacked MLPs, as shown in Fig.\,\ref{figure1}. The convolution-branch is composed of the Sampling and Grouping (SG) layer which is the same as that in PointNet++\cite{qi2017pointnet++} and two blocks of MLPs.

Assuming that the input local feature is $\mathbf{F}_{l}^{in}\in \mathbb{R}^{N_{in}\times C_{in}}$, where $N_{in}$ is the number of input points and $C_{in}$ is the number of feature channels, the convolution-branch process it as follows. Firstly, the SG layer samples and groups the $\mathbf{F}_{l}^{in}$ to $\mathbf{F}_1\in \mathbb{R}^{N_{out}\times S\times C_{in}}$, where the $N_{out}$ is the number of sampled points and $S$ is the number of grouped neighbor points.  The first block of MLPs layer embeds the $\mathbf{F}_1$ to $\mathbf{F}_2$, whose dimension is $N_{out}\times S\times C_2$.  Then, the second block of MLPs embeds the  $\mathbf{F}'_2$, which is the sum of local feature $\mathbf{F}_2$ and the transmitted global feature $ft_2\left( \mathbf{F}_g \right) $, to $\mathbf{F}_3$, whose dimension is $N_{out}\times S\times C_{out}$. Lastly, the maximum pooling is performed on the neighbor channel of $\mathbf{F}_3$ to aggregate the feature from neighbors and the output local feature $\mathbf{F}_{l}^{out}\in \mathbb{R}^{N_{out}\times C_{out}}$ is obtained.

The operation of two blocks of MLPs represents the local convolution operator $conv_{1}^{i}\left( \odot \right) $ and $conv_{2}^{i}\left( \odot \right) $ in Eq.\,(\ref{formular1}, where the first block of MLPs in Fig.\,\ref{figure1} implements operator $conv_{1}^{i}\left( \odot \right) $ and the second block implements operator $conv_{2}^{i}\left( \odot \right) $.

Performing convolution-like operation, which is implemented by the convolution-branch in the CT-block, on the grouped neighbor points can extract the local features of the point cloud, but this method loses the long-distance global information. The transformer-branch can obtain long-distance global information based on the attention mechanism, but it is hard to obtain the local details. In the proposed CT-block, the first block of MLPs on the convolution-branch can transmit the learned local feature to transformer-branch through the first feature transmission element. Meanwhile, the second block of MLPs on the convolution-branch can obtain the extracted global feature from the transformer-branch through the second feature transmission element. The proposed CT-block implements the interaction between the two branches through the feature transmission element, making the extracted features more expressive.

\begin{figure}[ht!]
	\centering\includegraphics[width=16cm]{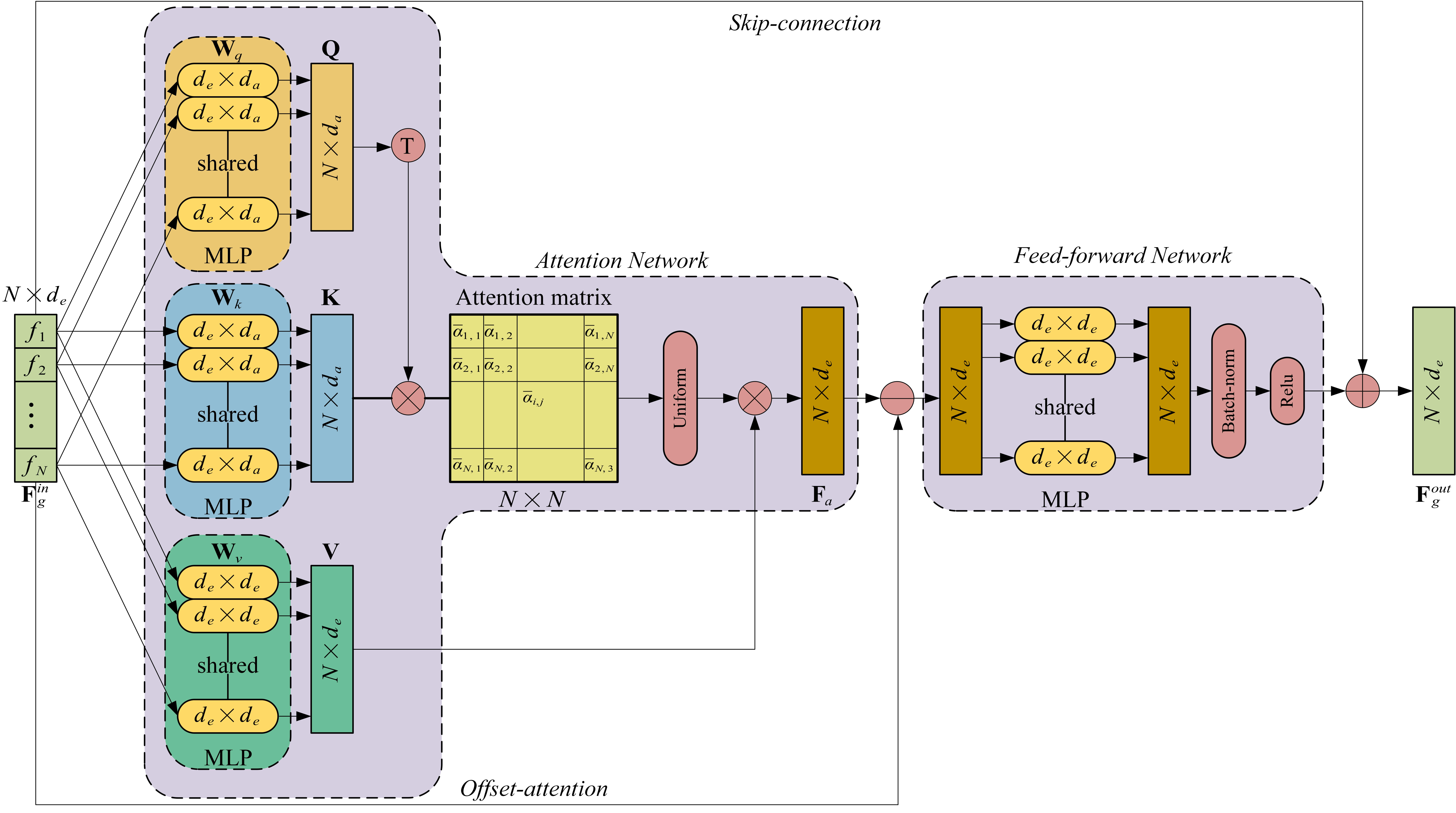}
	\caption{The structure of OA-module.}
	\label{figure2}
	\vspace{0cm}
\end{figure}

\textbf{Transformer-branch} based on the attention mechanism extracts the global feature of the point cloud. The point cloud is a kind of unordered data and attention-based transformer is suitable for extracting the global feature of unordered data\cite{guo2021pct}. 

The Offset-Attention (OA) mechanism is more suitable for processing point cloud than the Self-Attention (SA) mechanism\cite{guo2021pct}, the OA mechanism subtracts the input features from the self-attention features so that the coefficient matrix for multiplying the input features is changed from the adjacency matrix $\mathbf{E}$ to the Laplacian matrix $\mathbf{L}$\cite{guo2021pct}. And therefore, we use OA-mechanism to construct transformer-branch to learn the global feature of the point cloud. The specific structure of the OA-module is shown in Fig.\,\ref{figure2}. OA-module mainly consists of the following three steps.

The first step is to obtain the normalized attention matrix. Assuming that the input global feature of the transformer-branch is $\mathbf{F}_{g}^{in}\in \mathbb{R}^{N\times d_e}$, where $N$ is the number of input points and $d_e$ is the number of embedding feature channels, the \textit{query}, \textit{key}, and \textit{value} matrix denoted as $\mathbf{Q}$, $\mathbf{K}$, and $\mathbf{V}$ respectively, are obtained by the linear transformation of the input features according to the following equation\cite{vaswani2017attention}
\begin{equation}
\begin{aligned}
&\left( \mathbf{Q},\mathbf{K},\mathbf{V} \right) =\mathbf{F}_{g}^{in}\left( \mathbf{W}_q,\mathbf{W}_k,\mathbf{W}_v \right) \,\,\\
&\mathbf{Q},\mathbf{K}\in \mathbb{R}^{N\times d_a},\mathbf{V}\in \mathbb{R}^{N\times d_e}\\
&\mathbf{W}_q,\mathbf{W}_k\in \mathbb{R}^{d_e\times d_a},\mathbf{W}_v\in \mathbb{R}^{d_e\times d_e}
\end{aligned}
\label{formular2}
\end{equation}
where $\mathbf{W}_q$, $\mathbf{W}_k$, $\mathbf{W}_v$ are the learnable linear transformer layers; $d_a$ is the dimension of \textit{query} and \textit{key} vectors, and $d_e$ is the dimension of \textit{value} vector. After the \textit{query} matrix $\mathbf{Q}$ is multiplied by the transposed \textit{key} matrix $\mathbf{K}^\mathrm{T}$, the attention matrix can be obtained, as described in the following equation
\begin{equation}
\mathbf{\bar{A}}=\left( \bar{\alpha} \right) _{i,j}=\mathbf{Q}\cdot \mathbf{K}^{\mathrm{T}}
\label{formular3}
\end{equation}
where $\mathbf{\bar{A}}\in \mathbb{R}^{N\times N}$ is the attention matrix. For the stability of the calculation, the attention matrix needs to be normalized\cite{vaswani2017attention}. The normalization process of the attention matrix of OA-mechanism is different from that of SA-mechanism and follows the equations\cite{guo2021pct}
\begin{equation}
\begin{aligned}
&\hat{\alpha}_{i,j}=\mathrm{soft}\max \left( \bar{\alpha}_{i,j} \right) =\frac{\exp \left( \bar{\alpha}_{i,j} \right)}{\sum_k{\exp \left( \bar{\alpha}_{k,j} \right)}}\\
&\alpha _{i,j}=\frac{\hat{\alpha}_{i,j}}{\sum_k{\hat{\alpha}_{i,k}}}
\end{aligned}
\label{formular4}
\end{equation}
where $\alpha _{i,j}$ is the normalized weight of the attention matrix. In the case that the input is point cloud data, the weight $\alpha _{i,j}$ can be regarded as the importance of the $j$-th point to the $i$-th point. 

The second step of the OA-module is to apply the normalized attention matrix to conduct the self-attention on the \textit{value} vector. The self-attention process makes the output features $\mathbf{F}_a\in \mathbb{R}^{N\times d_a}$ be the weighted sum of the \textit{value} vector $\mathbf{V}$, where the weight is from the uniformed attention matrix $\mathbf{A}$. The $\mathbf{F}_a$ can be calculated by the following equation,
\begin{equation}
\mathbf{F}_a=\mathbf{A}\cdot \mathbf{V}
\label{formular5}
\end{equation}

In the case that point cloud data, the $i$-th row of $\mathbf{F}_a$ represents the feature of the $i$-th point after the self-attention operation. The most important is that the $i$-th row feature is the weighted sum of \textit{value} features of all points according to the weight of the attention matrix, which follows the equation
\begin{equation}
\left( f_a \right) _{i}^{\mathrm{T}}=\sum_k{\alpha _{i,k}}v_{k}^{\mathrm{T}}
\label{formular6}
\end{equation}
where $\left( f_a \right) _{i}^{\mathrm{T}}\in \mathbb{R}^{1\times d_a}$ represents the $i$-th feature vector obtained by the self-attention operation; $v_{k}^{\mathrm{T}}\in \mathbb{R}^{1\times d_a}$ represents the \textit{value} feature of the $k$-th point; The weight $\alpha _{i,k}$ represents the importance of the $k$-th point to the $i$-th point. In other words, after the self-attention operation, the output features of each point are aggregated from the features of all points, that is, the self-attention operation has a global receptive field to extract global features.

The third step of the OA-module is to implement offset-attention. Firstly, input feature $\mathbf{F}_{g}^{in}$ is subtracted from the self-attention feature $\mathbf{F}_a$. Then, the subtracted feature is passed through a forward propagation network. Finally, the input feature $\mathbf{F}_{g}^{in}$ is added to the feature from the forward propagation network to implement the skip-connected\cite{he2016deep}. The above offset-attention process can be obtained by the following equation\cite{guo2021pct}
\begin{equation}
\mathbf{F}_{g}^{out}=\mathbf{LBR}\left( \mathbf{F}_a-\mathbf{F}_{g}^{in} \right) +\mathbf{F}_{g}^{in}
\label{formular7}
\end{equation}
where $\mathbf{LBR}$ represents the forward propagation network and is composed of linear layer, Batch-Normalization(BA) layer and ReLU-activation layer.

The above offset-attention process of the transformer-branch represents the operator $trans\left( \odot \right) $ in  Eq.\,(\ref{formular1}). In summary, the OA-attention-based transformer-branch can extract the global features of the point cloud. However, the OA-attention-based transformer lacks the priori of the local area, and it is weak to learn the local feature. In the proposed CT-block, the local features extracted by the convolution-branch are added to the input global features of the OA-module through the first feature transmission element, which makes up for the shortcomings of the transformer-branch's insufficient ability to extract local features.

\textbf{Feature transmission element} can effectively implement the interaction between the convolution-branch extracting local features and the transformer-branch extracting global features. It mainly solves the following three problems. Firstly, since the local features output by convolution-branch are down-sampled by the SG layer and the dimension of global features output by transformer-branch remains unchanged, the number of points of local features and global features are different in the same CT-block. Therefore, the alignment of the number of points needs to be considered when fusing the two types of features. The number of points of local and global features are assumed as $N_l$ and $N_g$ respectively.  The first feature transmission element achieves the alignment of points number by interpolating the local features, that is, interpolating $N_l$ number of local feature to the $N_g$ number. The up-interpolation of local features adopts the distance-based interpolation criterion, which is the same as the up-interpolation method in PointNet++\cite{qi2017pointnet++}. The second feature transmission element achieves the alignment of points number by down-sampling the global features, that is, sampling $N_l$ number of features from the $N_g$ number of global features. During down-sampling, the position of each selected global feature in the underlying Euclidean space is corresponding to that of the local feature. To sum up, the feature transmission element solves the problem of the alignment between the points number of local features and that of the global feature.

Secondly, in the same CT-block, the feature channels number of the local feature is different from that of the global feature. The feature channels number of local features is $C_l$, while that of the global features is $C_g$. The feature transmission element consists of a MLP layer and a BA layer, where the MLP layer is used for the alignment of feature channels number, and the BA layer is used for feature regularization. Therefore, the feature transmission element solves the problem of feature channel alignment between the local and global features.

Finally, even if the local feature and the global feature have the same feature channels number and they can be directly added together without the MLP to align channels number, the semantic gap between the two kinds of features will make the fusion difficult. Therefore, the feature transformation of the MLP must be used to compensate for the semantic gap between the two kinds of feature spaces. The local features extracted by the convolution-branch are transformed into global feature space by the first feature transmission element, and the global features extracted by the transformer-branch are transformed into local feature space by the second feature transmission element. The MLP layer in the feature transmission element is responsible for filling the gap between the two kinds of features.

In summary, the feature transmission element as the bridge implements the interaction between the convolution-branch and transformer-branch and makes the local and global features guide each other and merge effectively during learning. Besides this, through the feature transmission element, the number of points and feature channels between local and global features are aligned, and the semantic gap between the two types of features is filled.

\subsection{Network structure}
\label{section3.3}

\begin{figure}[ht!]
	\centering\includegraphics[width=16cm]{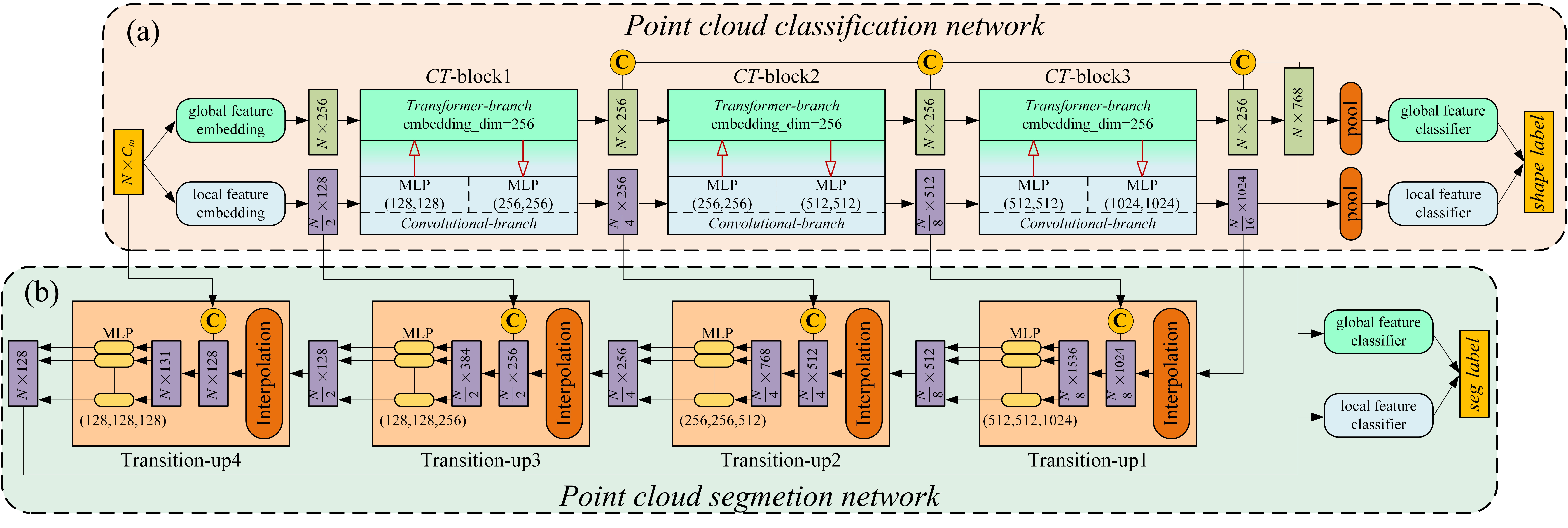}
	\caption{The structure of point cloud classification and segmentation networks constructed by the proposed CT-block. (a) The encoder of the classification network. (b) The decoder of the segmentation network}
	\label{figure3}
	\vspace{0cm}
\end{figure}

We use the CT-block as the feature extractor to construct the point cloud classification network and segmentation network, which are shown in Fig.\,\ref{figure3}.

\textbf{Point cloud classification network} is shown in Fig.\,\ref{figure3}(a). The input feature of the classification network is $\mathbf{F}_{in}\in \mathbb{R}^{N\times C}$, where $N$ is the points number of the input point cloud target, and $C$ is the feature channels number (if there is only position information, $C=3$. The feature channel can also contain information such as color, reflectivity, and surface normal. etc.). After the input feature $\mathbf{F}_{in}$ passes through two embedding blocks, the initial local feature and global feature are obtained. The local feature embedding block down-samples the input points number with a factor of 2 and increases the feature dimension, and the global feature embedding block keeps the original points number unchanged and increases feature dimensions. The local and global feature embedding blocks are both constructed by MLPs to embed the input point cloud with small number of feature channels to the initial features in high dimensional. The two types of initial features pass through three layers of CT-block. The convolution-branch of each CT-block down-samples the input points number with a factor of 2 and increase the feature dimension with a factor of 2, while the transformer-branch of each CT-block keeps the points number and feature dimensions unchanged. Finally, the two types of features are input into two classifiers respectively. The local features output by the last layer of CT-block are input to the local feature classifier after maximum pooling. The global features output by the three CT-blocks are firstly concatenated on the feature channel, and then the concatenated feature is input into the global feature classifier after maximum pooling. The output dimension of each classifier is $N_c$, which is the number of categories. We adopt the two-head classification strategy that the local and global feature classifiers are separately supervised by two cross entropy losses during training and the outputs of them are simply summarized with equal weight as the prediction results during inference.

\textbf{Point cloud segmentation network} is shown in Fig.\,\ref{figure3}(b). For dense prediction tasks such as the point cloud segmentation, we adopt a U-net\cite{ronneberger2015u} design in which the encoder is coupled with a symmetric decoder. The encoder of the segmentation network is the same as that of the classification network. In the decoder part, the local feature output by CT-block will be gradually restored to the original number of dense points through distance-based interpolation in the transition-up block, which is consistent with the up-interpolation method of PointNet++\cite{qi2017pointnet++}. The global features output by CT-block are not undergoing any processing. As in the point cloud classification network, the global and local features are also input into two classifiers respectively in the point cloud segmentation network. The local features output by the last transition-up layer of the decoder are input to the local feature classifier. The global features output by the three CT-blocks in the encoder are also firstly concatenated on the feature channel and then input into the global feature classifier. The output dimension of each classifier is $N\times N_s$, where $N$ is the original points number of target and $N_s$ is the number of segmentation categories.
During training and inference, we adopt the same two-head strategy as in the classification network. 

\section{Experiments}
\label{section4}
To evaluate the performance of the network composed by the proposed CT-block, we apply several widely used public datasets to conduct different task experiments. For the point cloud target classification task, we use the ModelNet40\cite{yi2016scalable} dataset. For the point cloud object part segmentation task, we use the ShapeNetPart\cite{wu20153d} dataset.  
Besides this, we also perform multiple ablation experiments to verify the effectiveness of the components of the CT-block and to illustrate the selection of their important parameters.

\subsection{Classification On the ModelNet40}
\label{section4.1}
We apply the ModelNet40 dataset to evaluate the performance of the point cloud classification network, which is shown in Fig.\,\ref{figure3}(a). The ModelNet40\cite{yi2016scalable} dataset contains 12311 CAD models from 40 man-made object categories. We follow the official dataset division that 9843 models for training and 2468 models for testing. Following the configuration of PointNet\cite{qi2017pointnet}, we uniformly sample 1024 points from the CAD model to generate the point cloud target.

During training, we adopt the conventional data augmentation method for training the point cloud classification network. We augment the dataset by randomly rotating the target around the z-axis and adding Gaussian jitter to each point of the target, where the mean of the Gaussian jitter is zero and the standard deviation of it is 0.02.
We apply the SGD optimizer with the momentum set to 0.9. The initial learning rate is set to 0.001 with the cosine annealing schedule to adjust it every epoch.
\begin{table}
	\caption{Results on the ModelNet40 shape classification. The evaluation metrics are the mean accuracy within each category (mAcc) and the overall accuracy (OA) over all classes.}
	\centering
	\begin{tabular}{ccccc}
		\hline
		\textbf{Method} 			& \textbf{Input} 					& \textbf{Point Number}     & \textbf{mAcc}  	&	\textbf{OA} 		\\
		\hline
		MVCNN\cite{su2015multi}						& image 							&		-					&	-				&   90.1	\\
		3DShapeNet\cite{wu20153d}					& voxel								&		-					&	77.3			&   84.7	\\
		VoxNet\cite{maturana2015voxnet}				& voxel								&		-					&	83.0			&   85.9	\\
		PointNet\cite{qi2017pointnet}				& point(\textit{xyz})				&		1k					&	86.2			&   89.2	\\
		PointNet++\cite{qi2017pointnet++}			& point(\textit{xyz,normal})		&		1k					&	-				&   91.9	\\ 
		PointConv\cite{wu2019pointconv}				& point(\textit{xyz,normal})		&		1k					&	-				&   92.5	\\ 
		PointCNN\cite{li2018pointcnn}				& point(\textit{xyz})				&		1k					&	88.1			&   92.2	\\
		KPconv\cite{thomas2019kpconv}				& point(\textit{xyz})				&		7k					&	-				&   92.9	\\
		PointWeb\cite{zhao2019pointweb}				& point(\textit{xyz})				&		1k					&	89.4			&   92.3	\\
		DGCNN\cite{wang2019dynamic}					& point(\textit{xyz})				&		1k					&	90.2			&   92.2	\\
		RS-CNN\cite{liu2019relation}				& point(\textit{xyz})				&		1k					&	-				&   92.9	\\
		PCT\cite{guo2021pct}						& point(\textit{xyz})				&		1k					&	-				&   93.2	\\
		\hline
		\textbf{Ours}								& point(\textit{xyz})				&		1k					&	\textbf{90.8}	&   \textbf{93.5}	\\
		\hline
	\end{tabular}
	\label{table1}
\end{table}

The experimental results of classification accuracy on the ModelNet40 test set are shown in Table\ref{table1}. From Table\ref{table1} we can find that the point cloud classification network composed of the proposed CT-block performs better than the PointNet++\cite{qi2017pointnet++}, which only extracts the local feature. The classification accuracy of the proposed network is higher than that of the PointNet++ by 1.6\%. Meanwhile, the proposed network performs better than the PCT\cite{guo2021pct}, which extracts local features and global features sequentially. The classification accuracy of the proposed network is higher than that of the PCT by 0.3\%. 
The experimental results show that because CT-block can simultaneously fuse and extract the local and global features of the point cloud, the point cloud classification network composed of CT-block has better performance.

\subsection{Object Part Segmentation On the ShapeNetPart}
\label{section4.2}
Point cloud object part segmentation is a fine-grained recognition task, which is to assign part category label (e.g. chair leg, cup handle) to each point of the input point cloud object.
We apply the ShapeNetPart dataset to evaluate the point cloud object part segmentation network, which is shown in Fig.\,\ref{figure3}(b). The ShapeNetPart\cite{wu20153d} dataset contains 16881 shapes from 16 categories annotated with 50 parts in total, and most of the categories are labeled with 2 to 4 parts. The official division of the ShapeNetPart dataset contains 14006 shapes for training and 2874 shapes for testing.

We follow the experimental configuration of PointNet\cite{qi2017pointnet}, where regarding part segmentation task as the classification of each point in the case of a known shape category and sampling 2048 points from each shape to generate the point cloud. 
During training, we only adopt the multi-scale data augmentation method, which augments the dataset by randomly anisotropic scaling the point cloud object, where the scaling ratio follows the uniform distribution [0.8,1.25]. We apply the SGD optimizer with momentum set to 0.9. The initial learning rate is set to 0.001 with a cosine annealing schedule to adjust it every epoch. During testing, we use two test strategies. The first is the single-scale test strategy, in which the input point cloud target maintains its original size. The second is the multi-scale test strategy which is the same in \cite{guo2021pct}, where the scale of the input point cloud target ranges from [0.8,1.25] with a step of 0.1.

\begin{table}[]
	\caption{Results on ShapeNetPart shape segmentation. The evaluation metrics are the part-average Intersection-over-Union (pIoU) and IoU for each categories.}
	\centering
	\resizebox{\textwidth}{!}
	{
	\begin{tabular}{c|c|cccccccccccccccc}
		\hline
		\textbf{Method}  & \textbf{pIoU}  & airplane & bag  & cap  & car  & chair & ear phone & guitar & knife & lamp & laptop & motor bike & mug  & pistol & rocket & skate board & 
		table \\
		\hline
		PointNet\cite{qi2017pointnet}            & 83.7 & 83.4      & 78.7 & 82.5 & 74.9 & 89.6  & 73.0      & 91.5   & 85.9  & 80.8 & 95.3   & 65.2       & 93.0 & 81.2   & 57.9   & 72.8        & 80.6  \\
		PointNet++\cite{qi2017pointnet++}        & 85.1 & 82.4      & 79.0 & 87.7 & 77.3 & 90.8  & 71.8      & 91.0   & 85.9  & 83.7 & 95.3   & 71.6       & 94.1 & 81.3   & 58.7   & 76.4        & 82.6  \\
		Kd-net\cite{klokov2017escape}        	 & 82.3 & 80.1      & 74.6 & 74.3 & 70.3 & 88.6  & 73.5      & 90.2   & 87.2  & 81.0 & 94.9   & 57.4       & 86.7 & 78.1   & 51.8   & 69.9        & 80.3  \\
		SSCNN\cite{yi2017syncspeccnn}          	 & 84.7 & 81.6      & 81.7 & 81.9 & 75.2 & 90.2  & 74.9      & 93.0   & 86.1  & 84.7 & 95.6   & 66.7       & 92.7 & 81.6   & 60.6   & 82.9        & 82.1  \\
		SpiderCNN\cite{xu2018spidercnn}      	 & 85.3 & 83.5      & 81.0 & 87.2 & 77.5 & 90.7  & 76.8      & 91.1   & 87.3  & 83.3 & 95.8   & 70.2       & 93.5 & 82.7   & 59.7   & 75.8        & 82.8  \\
		DGCNN\cite{wang2019dynamic}              & 85.2 & 84.0      & 83.4 & 86.7 & 77.8 & 90.6  & 74.7      & 91.2   & 87.5  & 82.8 & 95.7   & 66.3       & 94.9 & 81.1   & 63.5   & 74.5        & 82.6  \\
		PointConv\cite{wu2019pointconv}          & 85.7 & -         & -    & -    & -    & -     & -         & -      & -     & -    & -      & -          & -    & -      & -      & -           & -     \\
		PointCNN\cite{li2018pointcnn}           & 86.1 & 84.1      & 86.5 & 86.0 & 80.8 & 90.6  & 79.7      & 92.3   & 88.4  & 85.3 & 96.1   & 77.2       & 95.3 & 84.2   & 64.2   & 80.0        & 83.0  \\
		PointASNL\cite{yan2020pointasnl}         & 86.1 & 84.1      & 84.7 & 87.9 & 79.7 & 92.2  & 73.7      & 91.0   & 87.2  & 84.2 & 95.8   & 74.4       & 95.2 & 81.0   & 63.0   & 76.3        & 83.2  \\
		RS-CNN\cite{liu2019relation}             & 86.2 & 83.5      & 84.8 & 88.8 & 79.6 & 91.2  & 81.1      & 91.6   & 88.4  & 86.0 & 96.0   & 73.7       & 94.1 & 83.4   & 60.5   & 77.7        & 83.6  \\
		PCT(single-scale)                    	 & 86.1 & 83.8      & 85.4 & 87.7 & 80.6 & 91.3  & 72.2      & 91.0   & 87.4  & 84.7 & 96.0   & 71.8       & 95.3 & 82.4   & 61.7   & 74.6        & 83.8  \\
		PCT(multi-scale)\cite{guo2021pct}    	 & 86.4 & 85.0      & 82.4 & 89.0 & 81.2 & 91.9  & 71.5      & 91.3   & 88.1  & 86.3 & 95.8   & 64.6       & 95.8 & 83.6   & 62.2   & 77.6        & 83.7  \\
		\hline
\textbf{Ours}(single-scale)      				& 86.3 & 84.5      & 85.2 & 88.7 & 80.5 & 91.5  & 74.7      & 91.8   & 87.5  & 85.2 & 96.1   & 73.4       & 95.5 & 83.9   & 61.6   & 75.6        & 83.3  \\
\textbf{Ours}(multi -scale)      				& 86.5 & 84.7      & 84.4 & 88.1 & 81.0 & 91.8  & 75.9      & 92.2   & 87.8  & 85.6 & 96.0   & 73.7       & 95.3 & 83.9   & 59.7   & 74.9        & 83.6  \\
	\hline
	\end{tabular}
	}
	\label{table2}
\end{table}

In Table\ref{table2}, we use the part-average Intersection-over-Union (pIoU) and overall IoU for each shape category as metrics to evaluate our method, and these metrics are also for the PointNet++\cite{qi2017pointnet++}, PCT\cite{guo2021pct}, and other state of the art methods.
From Table\ref{table2} we can find that for most shape categories, the part segmentation network composed of the proposed CT-block performs better than PointNet++\cite{qi2017pointnet++}, which only extracts the local feature. Meanwhile, whether in the condition of the single-scale test or multi-scale test, the proposed network performs better than the PCT\cite{guo2021pct}, which extracts local features and global features sequentially.
The above experimental results show that CT-block can better extract and fuse global and local features, making the network composed of it performs better on the fine-grained part-segmentation task than networks that only extract local features or global features. 

\begin{figure}[ht!]
	\centering\includegraphics[width=16cm]{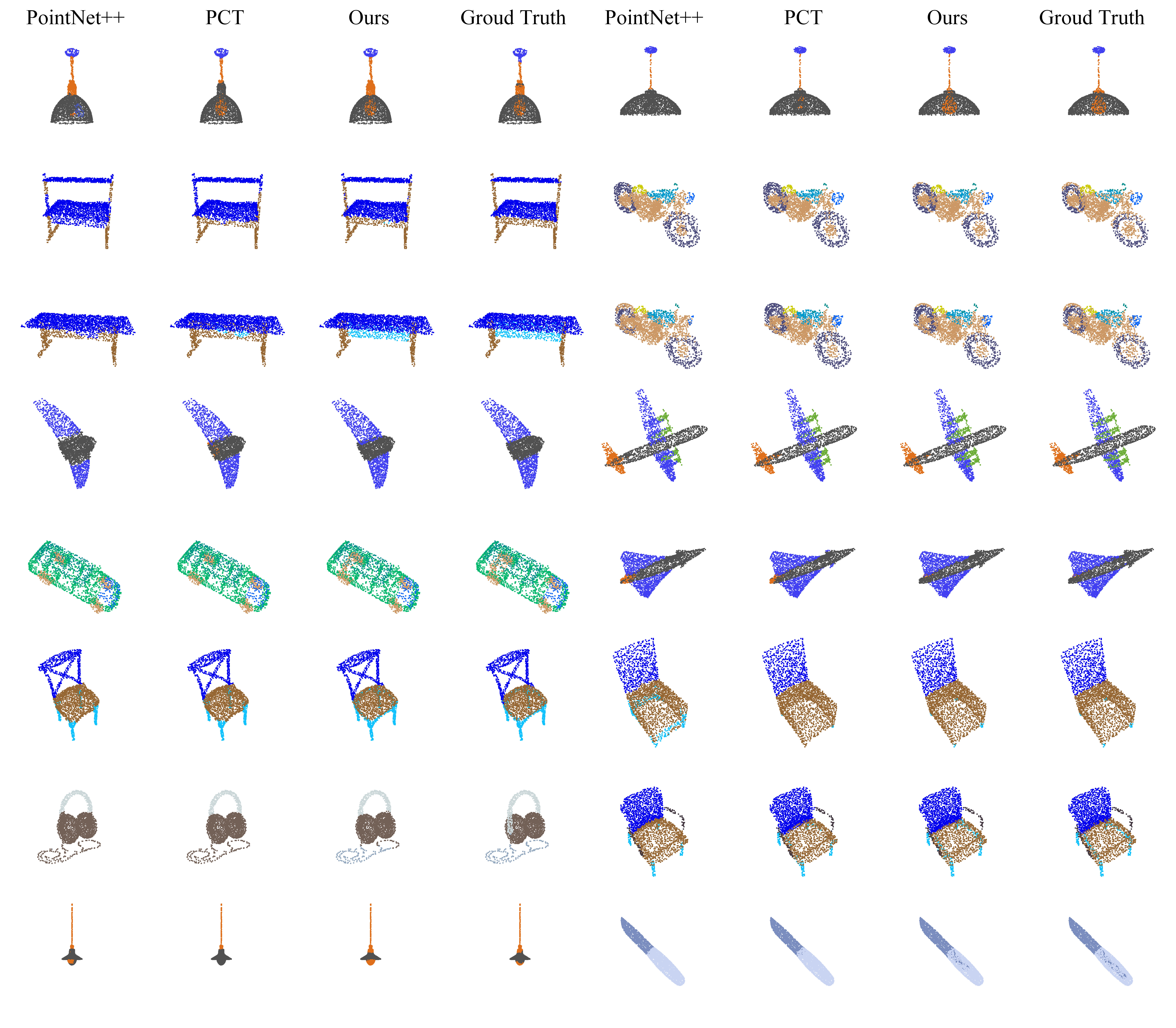}
	\caption{Object Part Segmentation reults from PointNet++, PCT, Ours method, and Ground Truth.}
	\label{figure4}
	\vspace{0cm}
\end{figure}

Fig.\,\ref{figure4} shows the segmentation results of PointNet++\cite{qi2017pointnet++}, PCT\cite{guo2021pct}, and our method. It can be seen from Fig.\,\ref{figure4} that the prediction of the proposed network has clearer details, such as the engine and wing of aircraft, the hole of table, the bulb of lamp, the leg of chair, etc, and is closer to the ground truth.

\subsection{Ablation Experiments}
\label{section4.4}
\subsubsection{The validity of the components of CT-block}
\label{section4.4.1}
The CT-block is composed of the convolution-branch, transformer-branch and feature transmission element. A single convolution-branch or transformer-branch is competent for the point cloud classification and segmentation tasks, while combining them with the feature transmission element can lead to better performance. In order to verify that the combination of the above three components can bring better performance, we construct the following four networks. The first network is only composed of the convolution-branch of CT-block, while the second network is only composed of the transformer-branch. The third network contains both convolution-branch and transformer-branch, but it doesn’t apply the feature transmission element as the bridge to fuse the learned feature. The last network is constructed by the complete CT-block, where the above three components work together. We evaluate the above four networks on the ModelNet40, ShapePartNet, 
and their performance is listed in the Table\ref{table4}. 

\begin{table}[]
	\caption{The performance of networks constructed by different part of CT-block on two datasets.}
	\centering
	\resizebox{\textwidth}{!}
	{
	\begin{tabular}{c|ccc|cc}
		\hline
		Network & Convolution-branch & Transforme-branch & Feature transmission element  & Accuracy on ModelNet40   & pIoU on ShapNetPart    \\
		\hline
		1st     & $\surd$                  & $\times$    & $\times$                      & 91.82                     & 85.23                     \\
		2nd     & $\times$                 & $\surd$     & $\times$                      & 91.75                     & 85.51                      \\
		3rd     & $\surd$                  & $\surd$     & $\times$                      & 92.59                     & 85.70                      \\
		4th     & $\surd$                  & $\surd$     & $\surd$                       & 93.52                     & 86.29                      \\
		\hline
	\end{tabular}
	}
	\label{table4}
\end{table}

From Table\ref{table4} we can find that, the classification accuracy of the fourth network is higher than that of the other three networks by 1.70\%, 1.77\%, 0.93\% respectively. The part segmentation accuracy of the fourth network is higher than that of the other three networks by 1.06\%, 0.78\%, 0.59\% respectively.  
For the three different tasks, the performance of the first network constructed by the convolution-branch that only extracts local features and the second network constructed by the transformer-branch that only extracts global features is worse than the third network that simultaneously extracts both kinds of features. However, the third network lacks the feature transmission element as the bridge to allow the two types of features to guide and merge with each other during learning, which leads to its performance being worse than the fourth network which is composed of the complete CT-block. The experimental result shows that each component of CT-block is significant for the performance of the network.

\subsubsection{The number of neighbors of the convolution-branch}
\label{section4.4.2}
When the convolution-branch of CT-block extracts local features from grouped neighbor points, the number of neighbor points $S$ will change the receptive field and the computational cost. The more number of grouped neighbor points, the larger the receptive field and computational cost. 
The proper number of neighbors on the convolution-branch is significant for the performance of the network. On the one hand, the transformer-branch has provided global information for convolution-branch through the feature transmission element in the CT-block, so the large number of neighbors with large receptive field may not bring much performance improvement but it will bring a huge computational cost. On the other hand, the small number of neighbors with small receptive fields may not be able to effectively extract local features and leading to bad performance. Therefore, we organized the experiment of selecting the number of grouped neighbor points to determine the reasonable scale for the local feature extraction.
We set the $S$ as 8, 16, 32, 48, 64,128 respectively in the point cloud classification network which is shown in Fig.\,\ref{figure3}. The model size(number of parameters (Params)), computational cost(floating point operations required (FLOPs)) of different networks and the classification accuracy on the ModelNet40 test set are shown in Table\ref{table5}.

\begin{table}[]
	\caption{The performance on the ModelNet40 and computation resource requirements of classification network with different grouped neighbor points number $S$ of the convolution-branch.}
	\centering
	\begin{tabular}{|c|c|c|c|}
	\hline
	Grouped neighbor Point number $S$ & \#Params(M)  & \#FLOPS(G)    & Accuracy \\
	\hline
	8                                 &    3.91      &    1.72       & 92.03     \\
	\hline
	16                                &    3.91      &    2.32       & 92.59     \\
	\hline
	32                                &    3.91      &    3.52       & 93.52     \\
	\hline
	64                                &    3.91      &    5.92       & 93.57    \\
	\hline    
	128                               &    3.91      &    10.73      & 93.45    \\
	\hline 
\end{tabular}
	\label{table5}
\end{table}

From Table\ref{table5} we can find that, (1) the number of model parameters does not change with the increase of $S$. (2) the computational cost increase with the increase of $S$. When the $S$ is too small, although the computational cost is low, the performance is bad. The large $S$ can improve the performance a little, but it brings much computational cost. To trade off the calculation and performance of the network, we recommend the number of neighbors of the convolution-branch in the CT-block as 32.

\subsubsection{The embedding dimension of the transformer-branch}
\label{section4.4.3}
The embedding dimension $d_e$ of transformer-branch affects the effectiveness of global feature extraction and changes the computational cost of network. In theory, the larger the embedding dimension, the greater the efficiency of extracting global features and the larger computational cost of network. We have to trade off the relationship between the effectiveness of global feature extraction and the computational cost of network.
Therefore, we organized the experiment of selecting the number of embedding dimension. We set the $d_e$ as 64,128, 256, 512 respectively in the point cloud classification network which is shown in Fig.\,\ref{figure3}. The model size, computational cost of different networks and the classification accuracy on the ModelNet40 test set are shown in Table\ref{table6}.

\begin{table}[]
	\caption{The performance on the ModelNet40 and computation resource requirements of the classification network with different embedding dimensions $d_e$ of the transformer-branch.}
	\centering
	\begin{tabular}{|c|c|c|c|}
	\hline
	Eembedding dimension  $d_e$       & \#Params(M) & \#FLOPS(G) & Accuracy \\
	\hline
	64                                 &   2.28       &   2.25      & 92.33     \\
	\hline
	128                                &   2.65       &   2.58      & 92.78     \\
	\hline
	256                                &   3.91       &   3.52      & 93.52     \\
	\hline
	512                                &   8.47       &   6.49      & 93.41    \\
	\hline    
	1024                               &   25.75      &  16.75      & 93.12    \\
	\hline 
\end{tabular}
	\label{table6}
\end{table}
From Table\ref{table6}, we can find that (1) the number of model parameters and the computational cost both increase with the increase of $d_e$. (2) When the $d_e$ is too small, although the computational cost is low, the performance of the model is bad. This means that the network with small $d_e$ can not effectively extract global features of the point cloud. However, when the $d_e$ is large, the performance improvement is not as much as the increase of computational requirements. And too large $d_e$ leads to the overfitting of the model. To ensure a reasonable amount of calculation and performance of the network, we recommend the number of embedding dimensions of the transformer-branch in the CT-block as 256.

\section{Conclusions}
\label{section5}
To make up for the shortcomings of convolution-like operation to extract global features and that of the transformer operation to extract local features, we propose a novel point cloud feature extractor CT-block which can simultaneously extract and merge the local and global features in this paper. The CT-block consists of a convolution-branch extracting local detailed features on the grouped neighbor points, a transformer-branch performing offset-attention on whole unordered point set to extract global features, and two feature transmission elements as the bridge to make two branches interact and guide each other during learning. We adopt the proposed CT-block to construct the point cloud classification and segmentation networks and compare their performance with some state-of-the-art methods on widely used datasets. The experiment results show that because the CT-block simultaneously extracts and merges the local and global features to get more expressive features, the point cloud classification and segmentation networks constructed by it achieve or perform better than some state of the art methods.

\section*{Acknowledgments}
This work was supported by the International Science and Technology Cooperation Project (No. 2015DFR10830). S. W. Guo thanks the National Science Foundation of China and International Science and Technology Cooperation Project for help identifying collaborators for this work.

\bibliographystyle{unsrt}  
\bibliography{references}

\end{document}